\documentclass{ieeeaccess}
\usepackage{cite}
\usepackage{amsmath,amssymb,amsfonts}
\usepackage{algorithmic}
\usepackage{graphicx}
\usepackage{textcomp}

\newcommand{\etal}{\textit{et al. }}
\usepackage{csquotes}

\def\BibTeX{{\rm B\kern-.05em{\sc i\kern-.025em b}\kern-.08em
    T\kern-.1667em\lower.7ex\hbox{E}\kern-.125emX}}
\begin{document}
\history{    }
\doi{   }

\title{Exploiting Global and Local Attentions for Heavy Rain Removal on Single Images}
\author{\uppercase{Dac Tung Vu}\authorrefmark{1}, 
\uppercase{Juan Luis Gonzalez}\authorrefmark{1}, 
\uppercase{Munchurl Kim}\authorrefmark{1}}

\address[1]{Korea Advanced Institute of Science and Technology, Daejeon, Korea}

\tfootnote{This work was supported by a grant from the Institute for Information \& Communications Technology Promotion (IITP) funded by the Korean government (MSIT) (No. 2017-0-00419, Intelligent High Realistic Visual Processing for Smart Broadcasting Media). This was also partially supported by the BK21 (Brain Korea 21) Program.}


\corresp{Corresponding author: Munchurl Kim (e-mail: mkimee@kaist.ac.kr).}

\begin{abstract}
Heavy rain removal from a single image is the task of simultaneously eliminating rain streaks and fog, which can dramatically degrade the quality of captured images. Most existing rain removal methods do not generalize well for the heavy rain case. In this work, we propose a novel network architecture consisting of three sub-networks to remove heavy rain from a single image without estimating rain streaks and fog separately. The first sub-net, a U-net-based architecture that incorporates our Spatial Channel Attention (SCA) blocks, extracts global features that provide sufficient contextual information needed to remove atmospheric distortions caused by rain and fog. The second sub-net learns the additive residues information, which is useful in removing rain streak artifacts via our proposed Residual Inception Modules (RIM). The third sub-net, the multiplicative sub-net, adopts our Channel-attentive Inception Modules (CIM) and learns the essential brighter local features which are not effectively extracted in the SCA and additive sub-nets by modulating the local pixel intensities in the derained images. Our three clean image results are then combined via an attentive blending block to generate the final clean image. Our method with SCA, RIM, and CIM significantly outperforms the previous state-of-the-art single-image deraining methods on the synthetic datasets, shows considerably cleaner and sharper derained estimates on the real image datasets. We present extensive experiments and ablation studies supporting each of our method's contributions on both synthetic and real image datasets.
\end{abstract}


\titlepgskip=-15pt

\maketitle

\section{Introduction}
\label{sec:introduction}

\PARstart{R}{ain} is a typical weather condition that affects the quality of images and videos, potentially degrading the performance of downstream tasks such as object tracking \cite{obj_tracking}, object detection \cite{obj_detection}, and autonomous driving \cite{auto_driving}. There are several works for rain removal on single images. However, most of them can only remove light rain efficiently and show very poor results for the heavy rain cases. Heavy rain is a much more challenging task as it entangles heavy rain streaks and haze/fog. A few works have addressed such a heavy rain removal problem, but still with lots of room for improvement.

Early rain removal methods applied hand-crafted priors \cite{GMM_2016, Luo_discriminative_sparse, Gu_JCAS_sparse, Deng_sparse, Zhu_JBO} such as sparse learning. Even when these methods improved the overall scene, they often resulted in artifacts in the derained images, such as blur and color distortions. Recently, with the development of the deep convolutional neural networks (DCNN), learning-based methods \cite{ Zhang_ID_CGAN, Li_RESCAN, Wang_SPANet, Ren_PreNet, Yasarla_UMRL, Zhang_DIDMDN} have shown dramatic improvements in deraining tasks with the help of large labeled datasets. However, these methods only achieve good results in the light rain removal task. For the heavy rain removal task, most approaches fail to restore the clean background scene.

Recently, Li \emph{et al.} \cite{Li_HeavyRain} proposed a heavy rain model as: 
\begin{equation}
\label{eq:ats}
    J = T \odot (I + S) + (1 - T) \odot A
\end{equation}
where {$J$} is the observed (heavy rain) image, {$I$} is the clean image, {$S$} is the rain streak layers, {$T$} is the transmission maps, and {$A$} is the atmospheric light of the scene. $\odot$ denotes the element-wise multiplication. This model considers rain streaks $S$ and fog ($T$ and $A$) but predicts $S$, $A$, and $T$ separately. In heavy rain, rain streaks and fog are mutually dependent by nature, making heavy rain removal a highly ill-posed problem. To tackle this issue,  Li \emph{et al.} \cite{Li_HeavyRain} applied a generative adversarial network (GAN) \cite{GAN_2014} to refine the background clean images. However, the brightness and color contrasts were not completely corrected. 

\begin{figure}
\begin{center}
\includegraphics[width=1\linewidth,keepaspectratio]{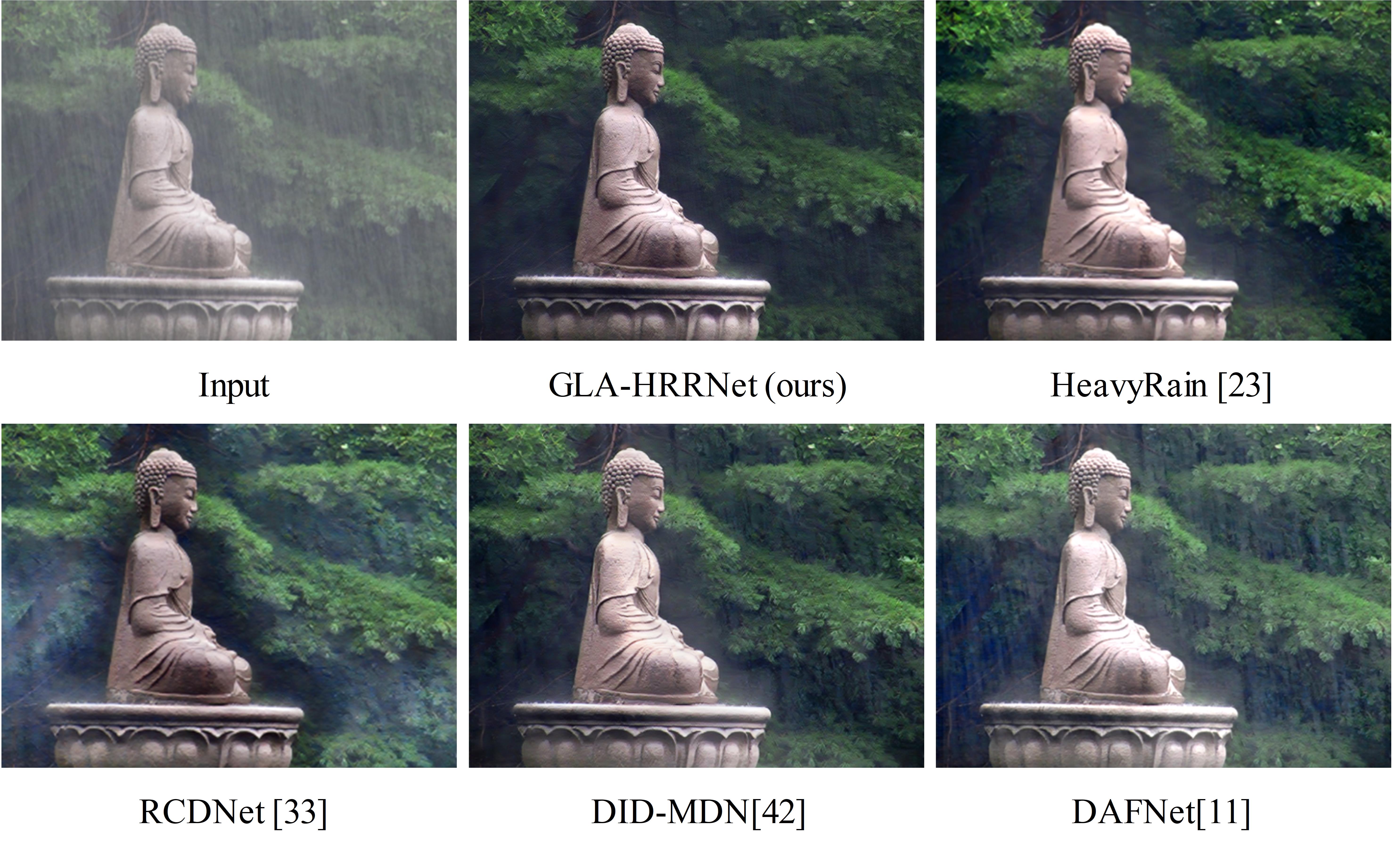}
\end{center}
\caption{Results on heavy rain removal for different SOTA methods. Our method generates a clean, derained, free-from-artifacts and realistic image. Meanwhile, the existing SOTA methods create an image with haze and incorrect color contrasts.}
\label{fig:first_fig}
\vspace*{-3mm}
\end{figure}

In another work, Hu \emph{et al.} \cite{Hu_DAF} considers the heavy rain scenes as a composition of a clean image $I$, a rain layer $R$, and a fog layer $F$ by formulating the heavy rain images $J$ as:
\begin{equation}
\begin{split}
\label{eq:daf}
J &= I\odot(1 - R - F) + R + AF \\
\end{split}
\end{equation}
Also, we re-arrange \eqref{eq:daf} for the clean image $I$ as:
\begin{equation} 
\label{eq:daf_res}
\begin{split}
I &= \dfrac{J+(-R-AF)}{1-R-F} \\
&= J\odot {\dfrac{1}{1-R-F}} +\dfrac{-R-AF}{{1-R-F}}
\end{split}
\end{equation}
where $0 \leq R \leq 1$ represents the rain layer, $A$ denotes an atmospheric light, and $0 \leq F \leq 1$ denotes the fog layer. In \cite{Hu_DAF}, to obtain the training pairs of clean and rainy images, rain layers $R$ and fog layers $F$ are computed by taking into account the scene depths. Hu \emph{et al.} \cite{Hu_DAF} proposed a residual network to learn a low-resolution depth map and depth-guided \textit{additive residues} for rain removal. But the clean image $I$ in \eqref{eq:daf_res} also have to be reconstructed based on the \textit{multiplicative residues} term with $J$, the first term in \eqref{eq:daf_res}, which could not be obtained from Hu \emph{et al.}'s residual network.  
In contrast with the previous methods \cite{Li_HeavyRain,Hu_DAF}, that indirectly generate the clean image by estimating the components in \eqref{eq:ats} and \eqref{eq:daf}, we directly generate the clean output image in our proposed method. In this paper, we propose to model the clean image $I$ as the combination of a roughly derained and defogged estimate $I_0$, an additive residue estimate $I_1$ (given by by $(J + R_A)$, where $R_A$ is the output of an additive sub-net), and a multiplicative residue estimate $I_2$ (which is provided by $(J \odot R_M)$, where $R_M$ is the output of a multiplicative sub-net), as defined by:
\begin{equation}\label{eq:GLA-HRR}
\begin{split}
    I &= I_0 \odot W_0 + (J + R_A) \odot W_1 + (J \odot R_M) \odot W_2 \\
     &= I_0 \odot W_0 + I_1 \odot W_1 + I_2 \odot W_2,
\end{split}
\end{equation}
where $W_0$, $W_1$ and $W_2$ blend pixel-wise the image components of $I$. 

In this paper, we propose a novel end-to-end network architecture for heavy-rain removal. Our network, called GLA-HRRNet, is integrated by three sub-nets that can effectively remove heavy rain: (i) In the first sub-net, we obtain global features from a U-net architecture where our proposed spatial channel attention (SCA) blocks are that incorporated, called the SCA sub-net; (ii) The second sub-net analyzes a heavy rain input into a clean image and additive residues with our proposed residual inception modules (RIM); (iii) The third sub-net employs the channel-attentive inception modules (CIM) to selectively learn the informative local features by separating the heavy rain scene into a clean image and multiplicative residues. Due to heavy rains' complex characteristics, some of the background information in the three sub-net outputs is estimated with degraded quality. To remedy this, we employed an attentive blending block to aggregate the three sub-nets' informative features into the final clean output image. Our main contributions in this paper are the following:
\begin{itemize}
    \item Firstly, we introduce an end-to-end network that simultaneously exploits global and local attentions to effectively remove heavy rain from single images, called GLA-HRRNet, which is a very challenging and ill-posed task in image restoration. 
    
    \item Secondly, we reinterpret the heavy rain model in terms of additive and multiplicative terms in our GLA-HRRNet, which effectively solves the heavy rain removal task without directly estimating the components in heavy rain scenes such as the atmospheric light, rain streaks, and transmission map. Our approach tackles the problem that there is no compatible formula to model heavy rain due to the entanglement of rain streaks and fog in natural scenes.
    
    \item Thirdly, we propose a spatial channel attention block that is effectively integrated into our Spatial Channel Attention (SCA) sub-net.
    
    \item Finally, we compare the proposed model with state-of-the-art (SOTA) deraining methods. Our method \textit{significantly} outperforms the SOTA methods in both synthetic and real image datasets by a considerable margin.
\end{itemize}

\section{Related Works}
In this section, we review related research in the field of image deraining. According to the type of input data, the existing works can be classified into two categories: video deraining methods and single image deraining methods.

\begin{figure*}
\begin{center}
\includegraphics[width=\textwidth,height=\textheight,keepaspectratio]{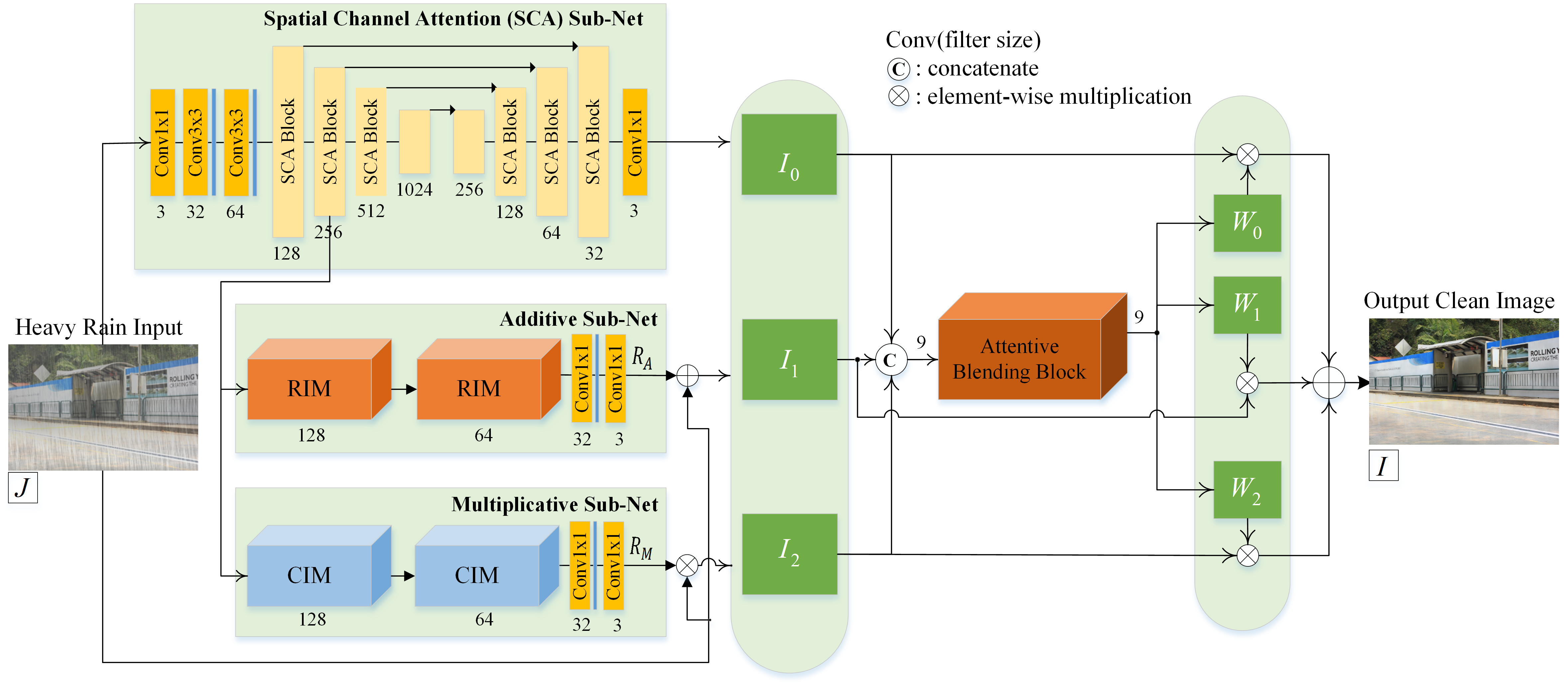}

\end{center}
\caption{The overall architecture of our Global and Local Attentive Heavy Rain Removal Network (GLA-HRRNet). The design of the Spatial channel attention (SCA) block is illustrated in Fig. \ref{fig:SCA-block}. The detailed architecture of the Residual inception modules (RIM) and Channel-attentive inception modules (CIM) is shown in Fig. \ref{fig:RIM-module} and Fig. \ref{fig:CIM-module}, respectively.} 
\label{fig:overall}
\end{figure*}

\subsection{Video Deraining}
The traditional video deraining methods process the relative information in sequential frames to detect and remove rain streaks. In \cite{Garg_Nayar_video_rain}, the authors proposed a correlation model that captures the dynamics of rain and a physics-based motion blur. In \cite{Kim_derain_desnow}, Kim \emph{et al.} obtained an initial rain map by subtracting warped frames from a current frame. They then remove the detected rain streaks by applying a low-rank matrix. On the other hand, the authors in \cite{Ren_matrix_decomposition} divide the rainy input images into sparse ones and dense ones and model them separately based on a matrix decomposition.

Moreover, with the dramatic development of DCNN \cite{Yang_Survey}, learning-based methods \cite{Progressive_rain_video}, \cite{Liu_erase_fill, Li_multiscale_sparse, Yang_self_learning} have significantly outperformed the traditional approaches. The work in \cite{Liu_erase_fill} firstly considers the rain removal problem, including rain occlusions. They remove rain from videos by constructing a recurrent DCNN to recover the spatial texture in rain occlusions guided by background reconstruction temporal coherence. Due to the repetitive local patterns and configuration characteristics of rain streaks, Li \etal \cite{Li_multiscale_sparse} proposed to apply a multi-scale convolutional model to remove rain streaks. Besides, Yang \etal \cite{Yang_self_learning} showed that the adjacent frames are highly correlated, and rain streaks are randomly distributed along the temporal dimension. Based on these two observations, they proposed a \enquote{self-learned} method consisting of two stages to remove rain streaks. This method was based on temporal correlation and consistency of frames in the video.

\subsection{Single Image Deraining}
Early rain removal methods for single images utilized hand-crafted priors such as Gaussian mixture models (GMMs) \cite{GMM_2016}, learning dictionaries and sparse architectures \cite{Luo_discriminative_sparse, Gu_JCAS_sparse, Deng_sparse}. Moreover, Zhu \emph{et al.} \cite{Zhu_JBO} proposed to compute the gradients in a local window to extract rainy regions, as rain streaks typically span in a narrow range of directions. The authors in \cite{Chen_low_rank} remove rain with the assumption that rain streaks have a similar direction for time interval. Although improving the overall scene, hand-crafted priors tended to generate artifacts such as blurring or distortion. 

In the deep learning era, learning-based methods \cite{Fu_DDN},\cite{Jiang_IADN, Fu_lightweight}, \cite{Wang_SPANet},\cite{Wang_fastderainnet}, \cite{Yang_JORDER},\cite{Yasarla_CMGR}, \cite{Zhang_ID_CGAN} have shown dramatic improvements from labeled datasets. Fu \emph{et al.} \cite{Fu_DDN} applied a low pass filter to decompose the rain image into two components, the base, and the detail parts, then utilized a ResNet for training their network to predict the detail component. Yang \emph{et al.} \cite{Yang_JORDER} employed multiple dilated convolutions to remove rain. Moreover, Zang \emph{et al.} \cite{Zhang_ID_CGAN} exploited a conditional GAN architecture, and perceptual loss \cite{Johnson_Perceptual} to train their model. Other authors have proposed using recurrent neural network (RNN) architectures to remove rain in multiple stages \cite{Li_RESCAN, Wang_SPANet}. These methods only achieve good results for light rain scenarios but fail to recover the background scene in heavy rain.

Relatively, less effort has been made on heavy rain removal. Li \emph{et al.} \cite{Li_HeavyRain} proposed a model to predict rain streaks, atmospheric light, and transmission maps, which are used to compute the initial clean image estimate. Additionally, Li \emph{et al.} refined the initial clean image estimate using a generative adversarial network (GAN). Wang \emph{et al.} \cite{Wang2020rethinking} applied a model considering a transmission map of rain streaks and fog. These methods do not efficiently remove heavy rain as their models struggle to represent the complex relations between heavy rain and fog in the single images. Different from the existing methods \cite{Li_HeavyRain, Hu_DAF, Wang2020rethinking}, we propose an end-to-end network to solve this problem.

In this work, we do not aim to model relative streaks, atmospheric lights, and transmission maps independently. Instead, we separate the heavy rain scene into a clean component, additive, and multiplicative residue components, which can better model the heavy rain imagery. More details are described in the next section.

\section{Methodology}

This section presents our proposed novel network architecture and loss functions for effective heavy rain removal in our method. Fig. \ref{fig:overall} shows our proposed overall architecture with three sub-nets and the attentive blending block used to combine informative features from these sub-nets to generate a final output image. We discuss the loss functions applied to the intermediate outputs and final results in Section \ref{sec:loss-func}.

\subsection{Network Architecture} \label{sec:network-arch}

\begin{figure}[t]
\begin{center}
\includegraphics[width=1\linewidth]{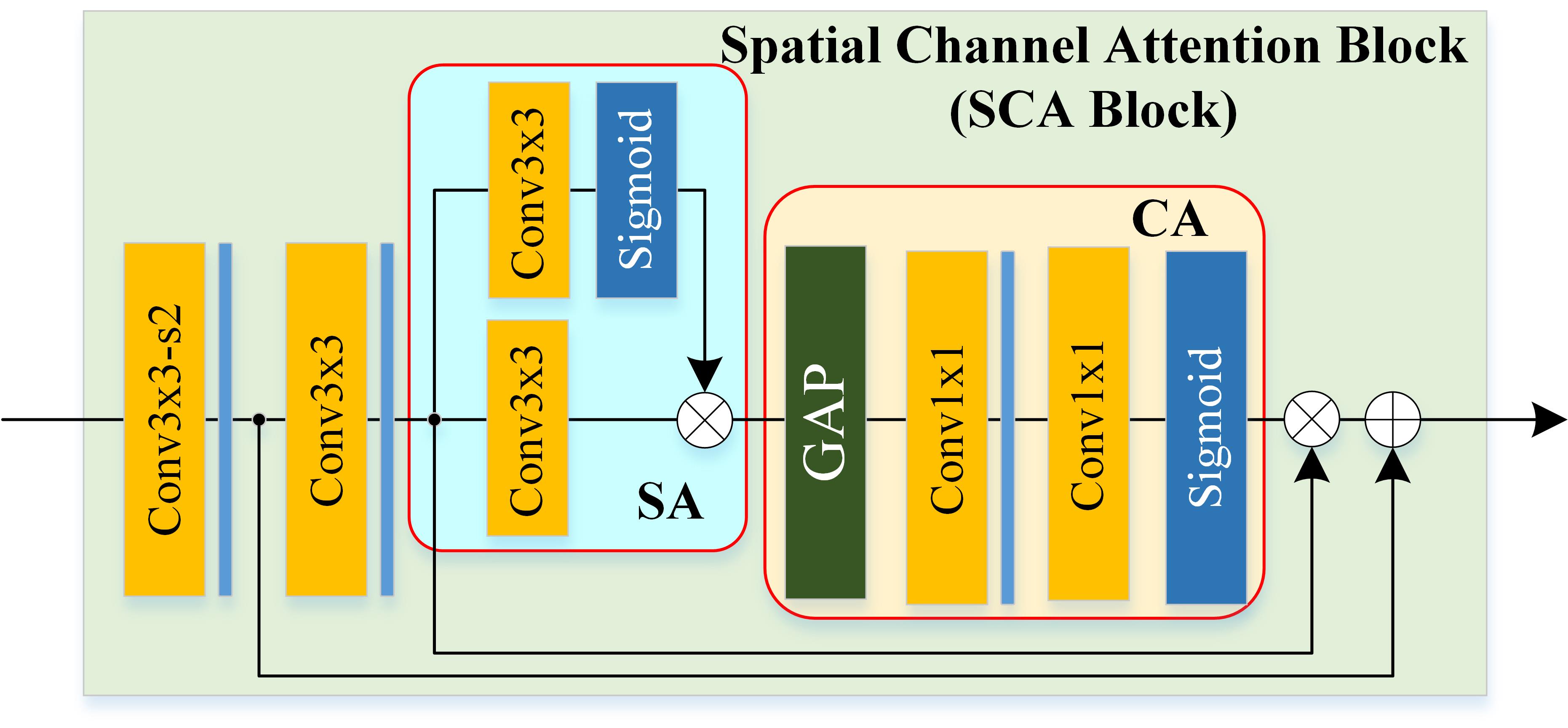}
\end{center}
   \caption{The architecture of Spatial Channel Attention Block (SCA Block). The first convolutional layer has a kernel size of 3 and a stride of 2. The following layers depict the Spatial Attention (SA) and the Channel Attention (CA) modules.}
\label{fig:SCA-block}
\end{figure}

Our network consists of three sub-nets: (i) a spatial channel attention (SCA) sub-net, (ii) an additive sub-net and (iii) a multiplicative sub-net, as shown in Fig. \ref{fig:overall}. The SCA sub-net takes heavy rain input images and directly yields rain-free image output. The SCA sub-net applies spatial channel attention (SCA) blocks based on a U-net architecture. The other two sub-nets predict residues from heavy rain scenes based on two operations: addition and multiplication. These two sub-nets take extracted features from the encoder part of the SCA sub-net as input. The second sub-net (additive sub-net) applies residual inception modules (RIM) to learn additive residues and adds additive residues to the input images to generate clean images. The third sub-net (multiplicative sub-net) employs channel-attentive inception modules (CIM) to produce multiplicative residues. After that, the multiplicative residues are multiplied by input heavy rainy images and generates clean images. However, some background information in the three clean image results is degraded because rain and fog are entangled and could not be easily separated. To solve this, we employ an attentive blending block to combine useful features from the three clean image estimates, $I_0, I_1, I_2$, and generate a final clean output image. We discuss more details about the three sub-nets in the following sections.

\subsubsection{Spatial Channel Attention (SCA) Sub-Net}
An auto-encoder based U-net \cite{U-net} is a specific architecture of a feedforward neural network with skip connections from its encoder part to the decoder part, which has been popularly used for image restorations. Our SCA sub-net is based on a U-net architecture that directly estimates a clean image from a heavy rain scene via learning a direct mapping.
The top left of Fig. \ref{fig:overall} depicts the SCA sub-net. The SCA sub-net has three convolution layers at the front to extract useful features from three-channel images. After that, the SCA blocks are followed as key components in the SCA sub-net. The combination of the SCA blocks in multi-scale resolutions helps the network learn global features to remove heavy rains effectively. Finally, one convolution layer constructs a three-channel clean image output. The SCA sub-net generates a coarse derained image ($I_0$) and extracts informative features ($F_{sca}$) that are later fed into the additive and multiplicative sub-nets, which are given by:
\begin{align} \label{eq:sca_net}
    I_0, F_{sca} = SCA\_subnet(J|\phi_{sca})
\end{align}
where $\phi_{sca}$ are the learnable filter parameters of our SCA sub-net.
 
The SCA block is illustrated in Fig. \ref{fig:SCA-block}. In this block, we first employ a convolution layer with stride 2 to downsample the input image in the encoder part (in the decoder part, this convolution layer is replaced with a transposed convolution layer to upsample the down-sized features). After that, we apply spatial attention (SA) (a light cyan block as shown in Fig. \ref{fig:SCA-block}) and followed by the channel attention (CA) (a flesh block in Fig. \ref{fig:SCA-block}). 
SA focuses on meaningful regions which contain rain streaks and fog in the heavy rain removal task. On the other hand, CA focuses on useful channels which are sensitive to rain streaks and fog. So, we combine both SA and CA to focus on informative regions and across channels to let the network efficiently remove heavy rain in single images with diverse shapes and directions of rain streaks and fog entanglement. The combination of CA and SA allows the SCA sub-net to be a main contributor to the reconstruction of the final output, even outperforming the SOTA methods as shown later in Section \ref{sec:experiments}.

\begin{figure*}
\begin{center}
\includegraphics[width=\textwidth,height=\textheight,keepaspectratio]{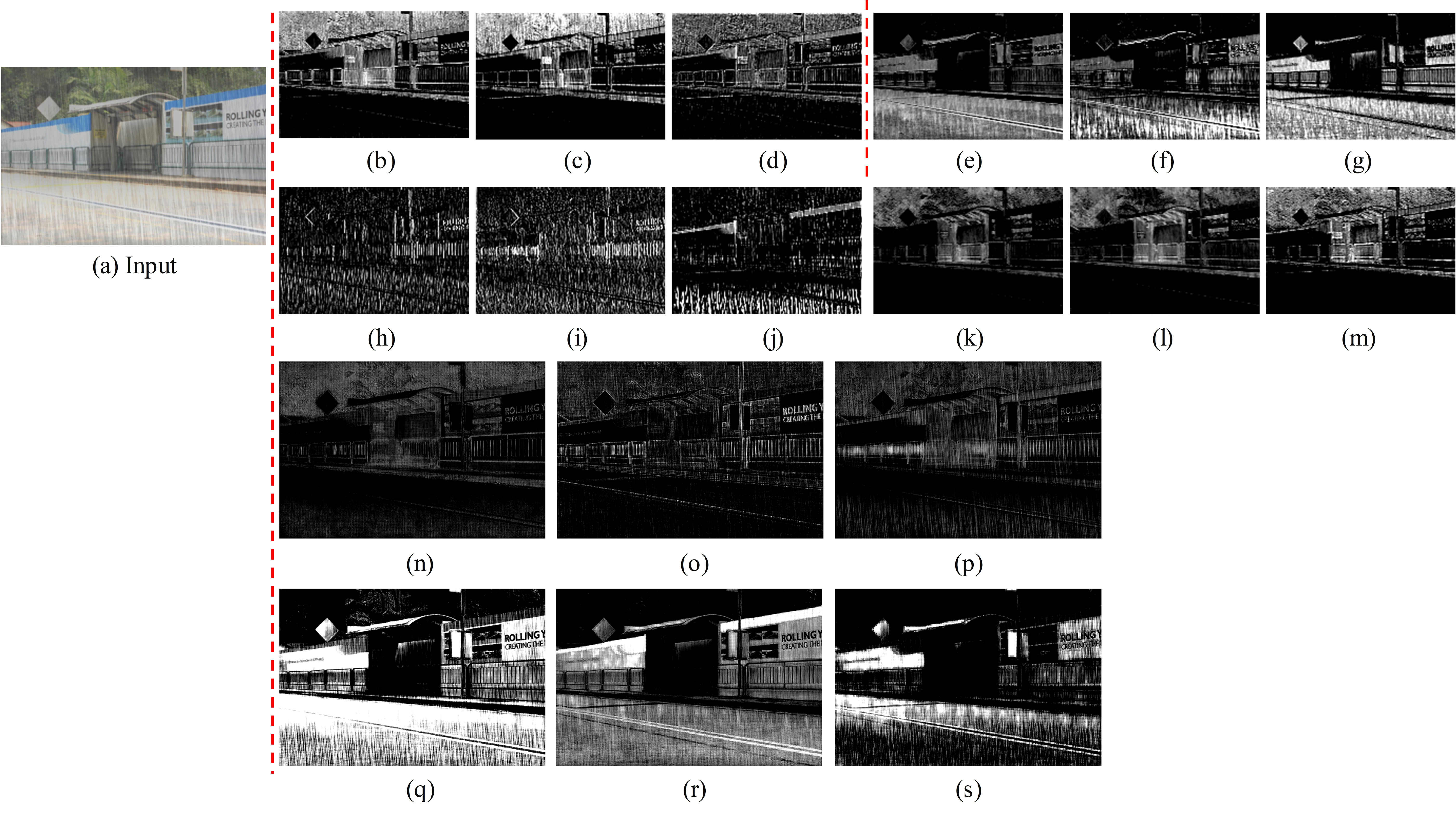}
\end{center}
\caption{Visualization of features extracted by SCA, Additive and Multiplicative sub-nets. The activation intensities for some interesting features maps of $F_{sca}$ are presented from (b) to (m). Three output features of the second RIM are illustrated from (n) to (p). And three output features of the second CIM are depicted from (q) to (s).} 
\label{fig:features}
\end{figure*}

Moreover, we extract informative features from lower resolutions in the encoder part of the SCA sub-net and feed them into the additive and multiplicative sub-nets. It was empirically found that passing the 256-channel features from the SCA sub-net, which are at 1/4 the input resolution, yields the best performance at a lower complexity than passing those of the layers with 512 and 1024 channels. Some of these extracted features are shown in Fig. \ref{fig:features}-(b) to Fig. \ref{fig:features}-(m). Especially, Fig. \ref{fig:features}-(a) presents heavy rain input, Fig. \ref{fig:features}-(b) to Fig. \ref{fig:features}-(d) illustrate features which are active in dark and detail regions where such active features are informatively explored in the following additive sub-net. Fig. \ref{fig:features}-(e) to Fig. \ref{fig:features}-(g) depict brighter regions which are explored for the multiplicative sub-net, and Fig. \ref{fig:features}-(h) to Fig. \ref{fig:features}-(m) exhibits the features that are active in rain streaks and fog in the input image. For more details, we analyze the roles of three sub-nets in Section \ref{sec:analysis_inter_results}.

\subsubsection{Additive Sub-Net}

\begin{figure}[t]
\begin{center}
\includegraphics[height=1.86in]{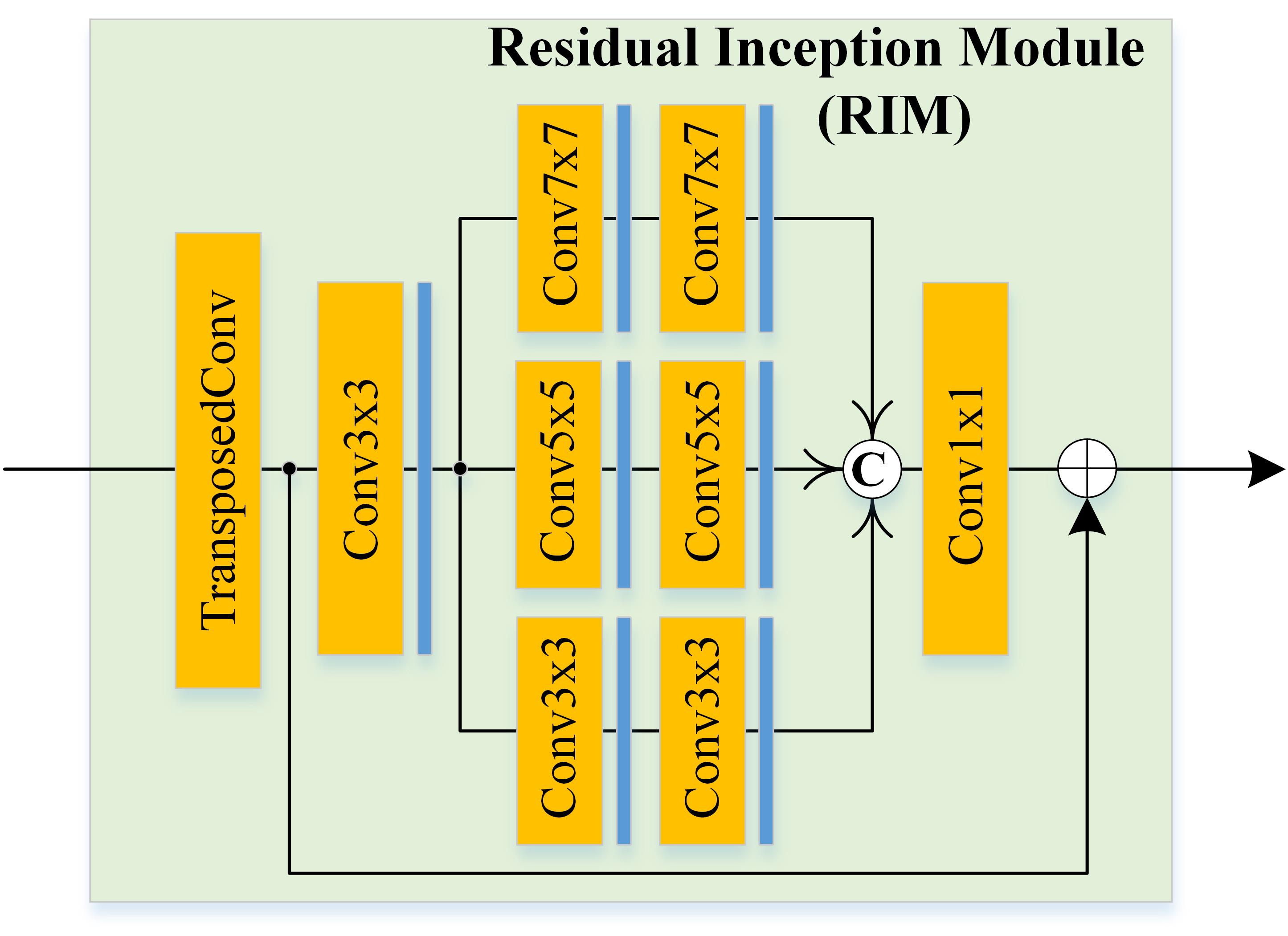}
\end{center}
   \caption{The architecture of our Residual inception module (RIM) in the additive sub-net.}
\label{fig:RIM-module}
\end{figure}

Removal of heavy rain in a single image is pretty complicated, so it is difficult to separate the physical components such as rain streaks, a transmission map, and atmospheric light from a heavy rain image. Therefore, to tackle this problem, we incorporate the additive and multiplicative sub-nets that separate the heavy rain image into a clean scene and residue components based on two operations: addition and multiplication. In the additive sub-net, we first extract features $(F_{sca})$ from the SCA sub-net's encoder and then incorporate two residual inception modules (RIM) to predict the additive residues. 
Fig. \ref{fig:features}-(n) to Fig. \ref{fig:features}-(p) visualize some output features of the second RIM. As can be observed, these features are more active in dark and detail regions, confirming that the additive sub-net helps recover high-frequency details and remove rain streaks in the dark regions.
After processing the SCA's features with two RIMs, two convolutional layers are applied to generate the additive residues for each of the three channels of the input images. The additive residues are combined with the heavy rain image input to generate an intermediate clean image estimate $I_1$ as followings:
\begin{align} \label{eq:addition}
    I_1(p) = J(p) + R_A(p)
\end{align}
where {$J$} is the heavy rainy input image, {$p$} is the pixel location, and $R_A$ denotes additive residues that are learned to decompose input {$J$} following \eqref{eq:GLA-HRR}. $ R_A$ is obtained from our additive sub-net by:
\begin{align} \label{eq:add_net}
    R_A = add\_subnet(F_{sca}|\phi_{add})
\end{align}
where $\phi_{add}$ are the learnable filter parameters of our additive sub-net. 

Fig. \ref{fig:RIM-module} shows the structure of the RIM used in our additive sub-net. This block, based on the inception block \cite{inception}, consists of three different convolution layers with kernel sizes of 3, 5, and 7, and then concatenates the results to learn additive residues. The three convolution layers with different kernel sizes allow for increased receptive fields and efficiently learn residues in heavy rain images with diverse shapes and various rain directions. With large receptive fields, the RIM block can effectively remove heavy rain streaks and artifacts that often appear in the output of the SCA sub-net.

\subsubsection{Multiplicative Sub-Net}

The deraining work \cite{Hu_DAF} that only employs additive residues learning has a limitation in removing heavy rains according to \eqref{eq:daf_res}. To remedy this shortage, we proposed a multiplicative sub-net to estimate multiplicative residue components. The multiplicative sub-net is incorporated with the additive sub-net and SCA sub-net to completely remove heavy rain from input images. Similar to the additive sub-net, the multiplicative sub-net firstly extracts features from the SCA sub-net. Two Channel-attentive Inception Modules (CIM) are incorporated to predict residues to the multiplication operation. Finally, convolutional layers estimate the multiplicative residues for each of the RGB channels. The clean image estimate $I_2$ of the multiplicative sub-net is given by:
\begin{align} \label{eq:multi}
    I_2(p) = J(p) \odot R_M(p)
\end{align}
where $R_M$ denotes the multiplicative residues in \eqref{eq:GLA-HRR}, and is obtained from our multiplicative sub-net by:
\begin{align} \label{eq:mul_net}
    R_M = mul\_subnet(F_{sca}|\phi_{mul})
\end{align}
where $\phi_{mul}$ are the learnable filter parameters of our multiplicative sub-net.

\begin{figure}[t]
\begin{center}
\includegraphics[height=1.86in]{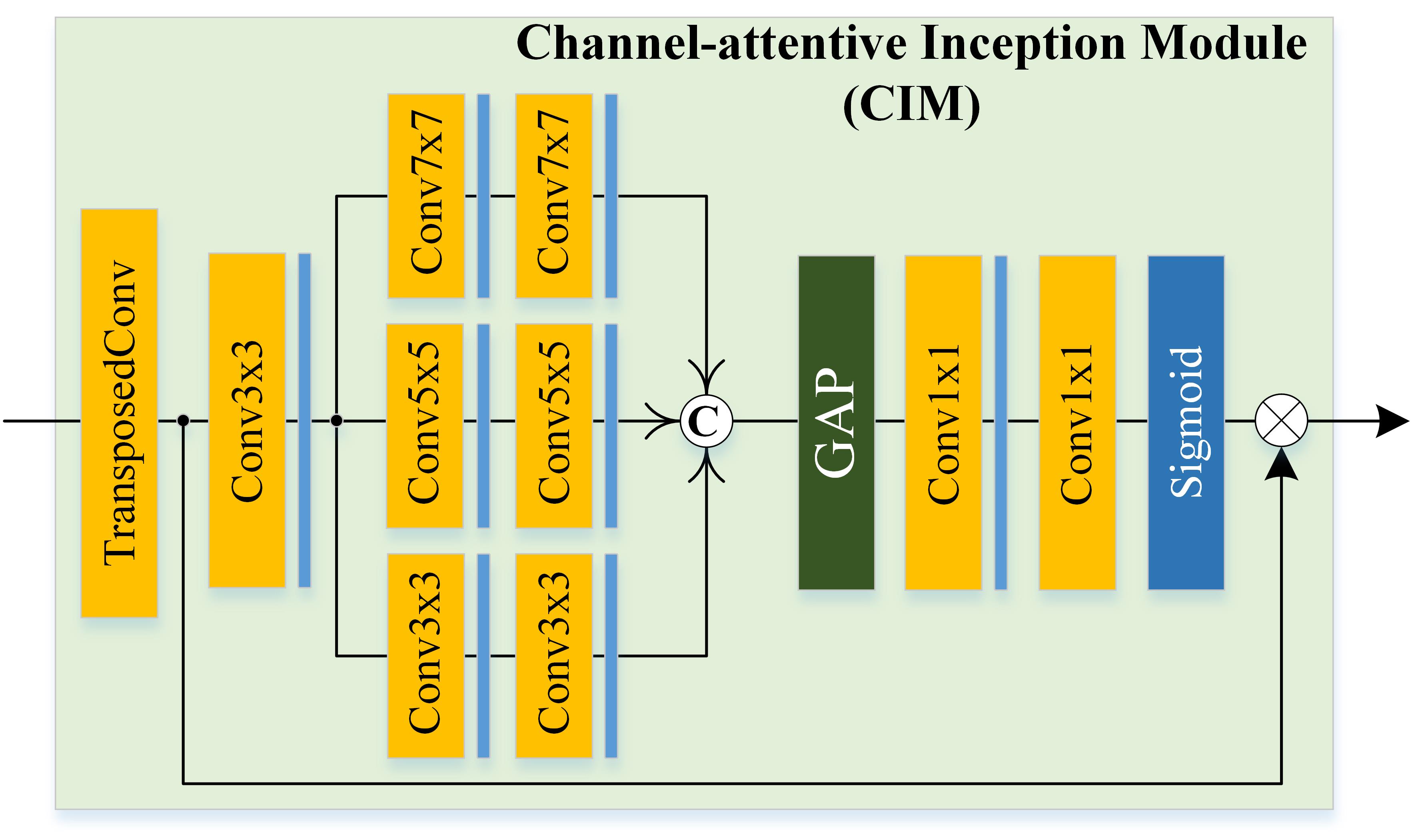}
\end{center}
   \caption{The architecture of our Channel-attentive Inception Module (CIM) in multiplicative sub-net.}
\label{fig:CIM-module}
\end{figure}

The multiplicative residues seem to be more complicated than the additive residues. These are also difficult to learn if a RIM architecture is used in the multiplicative sub-net. Hence, the proposed Channel-attentive Inception Module (CIM) is designed based on an inception block, followed by a global average pooling, two convolution layers, and a sigmoid function (channel attention) to efficiently learn the multiplicative residues. The cooperation between channel attention and inception helps focusing on more informative channels, making the multiplicative sub-net capable of learning the multiplicative residues. The architecture of the CIM is shown in Fig. \ref{fig:CIM-module}. Moreover, we depict some output features of the second CIM in Fig. \ref{fig:features}-(q) to Fig. \ref{fig:features}-(s). These features demonstrate that the multiplicative sub-net helps in modulating local pixel intensities in brighter regions.

\subsubsection{Intuition Behind Network Design}

\textbf{Why only Two Multiplicative and Additive Sub-nets:} The primary purpose of our separated models is the removal of rain streaks and fog entanglement that resides spread over the heavy rain images. The rain streaks are formulated as:
\begin{equation}
\label{eq:rain_streaks}
    J_s = I + S
\end{equation}
where $S$ indicates rain streaks, $J, I$ are the rainy input image and clean image, respectively. Besides, the fog image is modeled as:
\begin{equation}
\label{eq:fog}
    J_f = T \odot I + (1-T)\odot A
\end{equation}
where $T$ is transmission map, $A$ is atmospheric light and $J, I$ are the fog input image and clean output image, respectively. Meanwhile, the previous work \cite{Li_HeavyRain} separates the heavy rainy input image into the rain streaks, an atmospheric light, a transmission map, and background scene. The model is not effective because the rain streaks and fog are entangled in heavy rain scenes. Besides, the fog model and the rain streak model consist of only two arithmetic components: addition and multiplication according to \eqref{eq:rain_streaks} and \eqref{eq:fog}. Consequently, if generating heavy rain from rain streaks and fog in any formula, only two operators are employed: addition and multiplication. This has motivated us to propose two separator sub-nets (see Fig. \ref{fig:overall}) containing two components: the additive term {$J+R_A$} and the multiplicative term {$J*R_M$} in \eqref{eq:GLA-HRR}.

Moreover, to effectively combine the three clean images, which are provided by the SCA, additive, and multiplicative sub-nets, we use an attentive blending block to create weight maps that linearly combine the three clean estimates as the final rain-free output image. The weight maps are optimized when trained with a loss function of the network. Our superior performance shows the effectiveness of our sub-nets for heavy rain removal. More details of the attentive blending block are discussed in the following sub-section.

\begin{figure}[t]
\begin{center}
\includegraphics[width=1\linewidth]{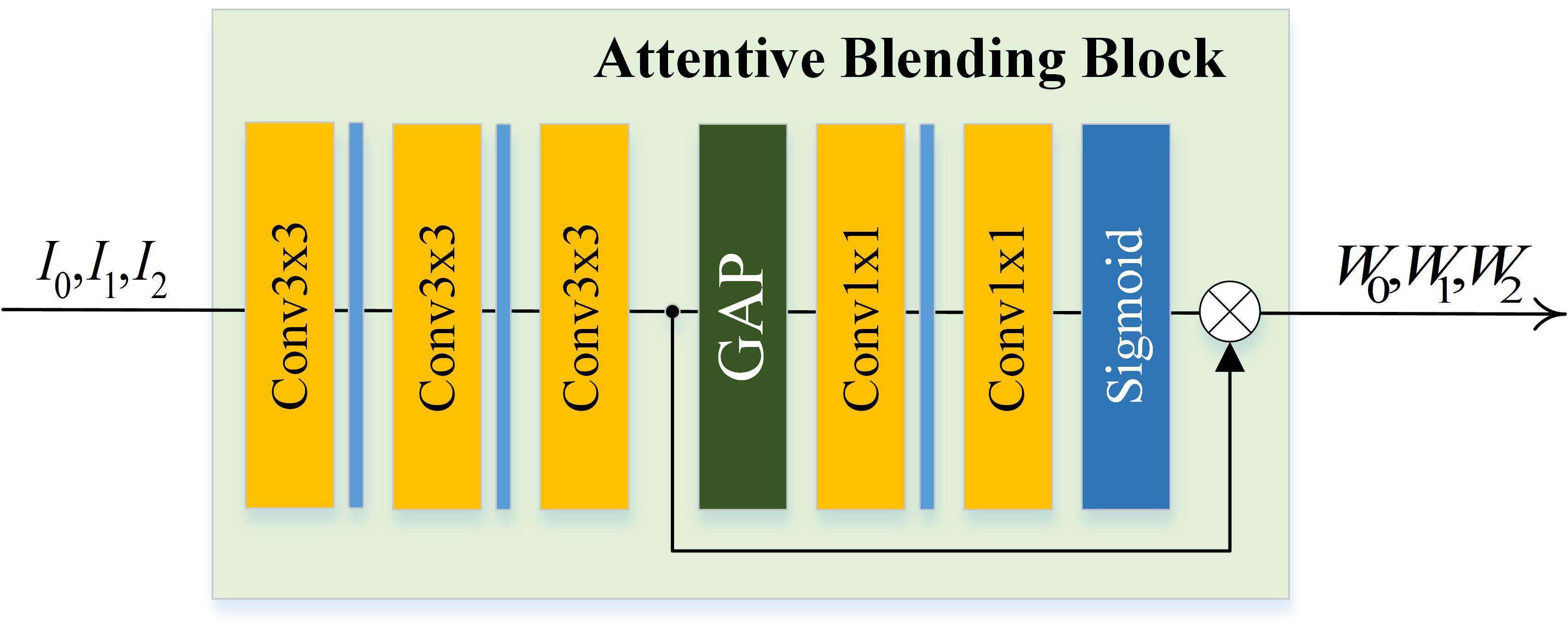}
\end{center}
   \caption{The architecture of the Attentive blending block.}
\label{fig:att-mechanism}
\end{figure}

\textbf{Attentive Blending Block:} After receiving the three clean estimates from the SCA, additive, and multiplicative sub-nets, we concatenate and feed them as input to the attention mechanism \cite{att_mechan_1, att_mechan_2} to estimate the final clean image. Expressly, three 3$\times$3 convolution layers, two 1$\times$1 convolution layers, a global average pooling layer, and a sigmoid function are incorporated. As expected, the background scene's information is automatically highlighted, leading to better recovery of a rain-free image. Furthermore, because there is complementary information in the three clean image results, the attentive blending block can selectively fuse the three intermediate outputs. Hence, our method combines these three sub-nets using these learned attention maps, thus outperforming the SOTA methods in image deraining. The architecture of the attentive blending block is shown in Fig. \ref{fig:att-mechanism}. The adaptive blending weights ($W_0$, $W_1$, and $W_2$) are then given by:
\begin{align} \label{eq:att_block}
    W_0, W_1, W_2 = ATT\_BB(I_0, I_1, I_2|\phi_{ATT\_BB})
\end{align}
where $\phi_{ATT\_BB}$ are the learnable filter parameters of our attentive blending block.

\subsection{Loss Functions} \label{sec:loss-func}

We apply an intermediate loss for the three clean image estimates to effectively learn useful features and efficiently remove heavy rain in the input images. As a result, the total loss used to train our end-to-end network architecture consists of two-loss functions measured at the intermediate and the final outputs, respectively. The intermediate loss includes three mean squared error (MSE) terms between the ground truth (GT) clean image and each of the clean estimates of the SCA, additive, and multiplicative sub-nets. The resulting intermediate loss ($L_{inter}$) is defined as follows:
\begin{align}
    L_{inter} &= \lambda_0L_{SCA}+\lambda_1L_{Add} + \lambda_2L_{Mul}\\
    L_{SCA} &= L2(I_{GT}, I_0)\\
    L_{Add} &= L2(I_{GT}, I_1)\\
    L_{Mul} &= L2(I_{GT}, I_2)
\end{align}
where $I_{GT}$ is the ground truth clean image. $I_0$, $I_1$, and $I_2$ are the clean estimates of the SCA, additive, and multiplicative sub-nets, respectively. $\lambda_0$, $\lambda_1$ and $\lambda_2$ are hyper-parameters that are determined empirically to have $\lambda_0 = \lambda_1 = \lambda_2 = 1$.

The loss function in the final output image includes the MSE loss and the edge loss between the clean image output and the GT as follows:
\begin{align}
    L_{final} &= \lambda_3L_{MSE}+\lambda_4L_{edge}\\
    L_{MSE} &= L2(I_{GT}, I)\\
    \begin{split}
    L_{edge} &= ||H_x(I_{GT})-H_x(I)||_1 \\
    &+ ||H_y(I_{GT})-H_y(I)||_1
    \end{split}
\end{align}
where $I$ is the final output of our GLA-HRRNet. {$H_x(\cdot), H_y(\cdot)$} are operators that calculate gradients along with the horizontal and vertical directions of the images, respectively. $\lambda_3$ and $\lambda_4$ are weighting hyper-parameters, empirically set to {$\lambda_3 = \lambda_4 = 1$}. The total loss function used to train our GLA-HRRNet can be expressed as:
\begin{align}
    L_{total} = L_{inter} + L_{final}
\end{align}

\begin{table} 
\setlength{\tabcolsep}{20pt}
\begin{center}
\begin{tabular}{l|l|l}
\hline
\hline
Method     & PSNR  & SSIM  \\ \hline
GLA-HRRNet (ours) & \textbf{25.58} & \textbf{0.880} \\ \hline
DAFNet \cite{Hu_DAF}           & 23.38 & 0.790 \\ \hline
HeavyRain \cite{Li_HeavyRain}   & 20.68 & 0.783 \\ \hline
SPANet \cite{Wang_SPANet}       & 19.12 & 0.746 \\ \hline
PreNet \cite{Ren_PreNet}        & 17.82 & 0.719 \\ \hline
RESCAN \cite{Li_RESCAN}         & 21.77 & 0.709 \\ \hline
DID-MDN \cite{Zhang_DIDMDN}     & 22.64 & 0.807 \\ \hline
UMRL \cite{Yasarla_UMRL}        & 21.58 & 0.816 \\ \hline
RCDNet \cite{Wang_RCDNet}       & 22.43 & 0.814 \\ \hline
MSPFN \cite{Jiang_MSPFN}        & 18.25 & 0.726 \\ \hline
JCAS    \cite{Gu_JCAS_sparse}   & 16.59 & 0.669 \\ \hline
\hline
\end{tabular}
\end{center}
\caption{The quantitative results of our proposed network and the state-of-the-art (SOTA) methods on the Outdoor-Rain dataset \cite{Li_HeavyRain}.}
\label{tab:result-syn-outdoor}
\end{table}

\section{Implementation Details} \label{sec:implementation}
\textbf{Training strategy:} First, we initialize the network parameters by the Xavier initialization \cite{Xavier-init}. Then, all training images are randomly cropped into patches with a size of 200$\times$300 pixels. We train our network for 200 epochs with a batch size of 4 using the Adam optimizer \cite{Adam} with the default settings. The learning rate is adjusted following the poly policy in \cite{Poly-policy} with an initial learning rate of 0.0001, which is halved after 100 epochs. We implemented our proposed GLA-HRRNet on PyTorch and performed our experiments on two NVIDIA TITAN X GPUs.

\textbf{Training datasets:} There are several large-scale synthetic datasets publicly available for training deraining models. However, most datasets do not contain the rain accumulation effects that are often present in the heavy rain scenarios. We selected the Outdoor-Rain dataset \cite{Li_HeavyRain} which is generated on a set of clean outdoor images. This dataset renders proper rain streaks and rain accumulation effects based on scene depths, which are estimated by a pre-trained monocular depth estimation \cite{monodepth}. The Outdoor-Rain dataset contains 9k training samples and 1.5k test samples.

\section{Experiments and Results}
\label{sec:experiments}

\begin{figure*}
\begin{center}
\includegraphics[width=\textwidth,height=\textheight,keepaspectratio]{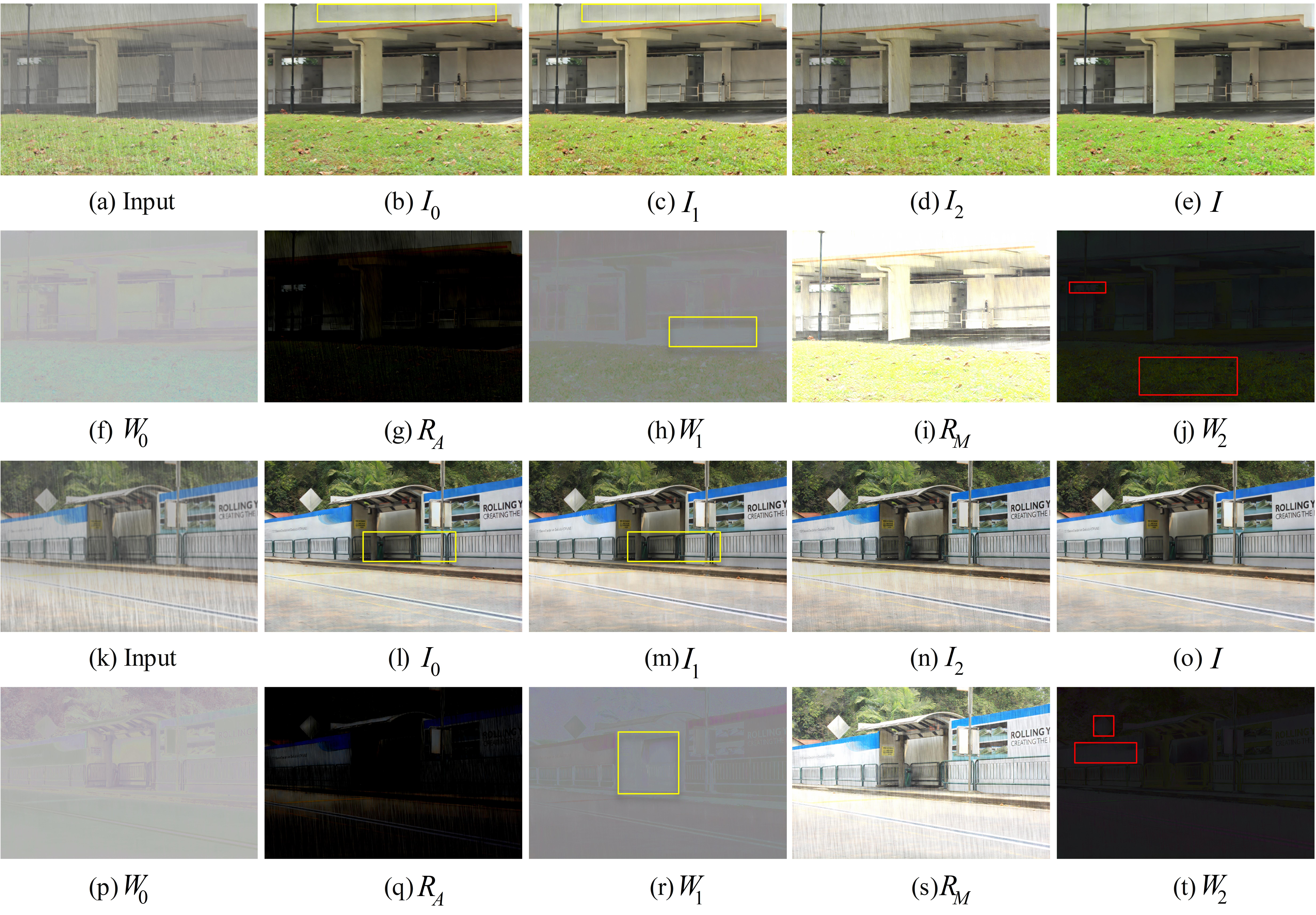}
\end{center}
\caption{Visualization of the components in our proposed GLA-HRRNet.}
\label{fig:visulization}
\end{figure*}

In this section, we compare our model, the GLA-HRRNet, with the state-of-the-art methods on both synthetic and real datasets. For the real dataset case, we take the real rainy images published in \cite{Li_HeavyRain}, and images collected from the internet with the keyword "heavy rain photo." The synthetic rain dataset is evaluated on the Outdoor-Rain dataset \cite{Li_HeavyRain} as mentioned in Section \ref{sec:implementation}. For a fair comparison, we download the publicly available source codes and re-train the previous works to generate clean image results from heavy rain images. If the SOTA networks do not converge, we use the best pre-trained models in the corresponding released source codes. 

\subsection{Analysis of Intermediate Results}
\label{sec:analysis_inter_results}
As discussed above, our proposed GLA-HRRNet integrates three intermediate sub-nets to generate a final clean output image. This section illustrates the three clean estimates obtained from the three sub-nets: SCA, additive, and multiplicative sub-nets. Fig. \ref{fig:visulization} shows three clean estimates, ($I_0, I_1, I_2$) and their corresponding weight maps ($W_0, W_1, W_2$) from the SCA, additive and multiplicative sub-nets, respectively. We also illustrate the output of additive sub-net, $R_A$ and multiplicative sub-net, $R_M$. Moreover, Fig. \ref{fig:visulization}-(e) and -(o) depict the final clean output images $I$.

Overall, we can observe that $I_0$ yields a better visual quality in terms of deraining than those of $I_1$ and $I_2$ for each input. This is because the SCA sub-net extracts global features, which help to capture scene geometry information. Nevertheless, it can be noted that $I_0$ still contains artifacts as some features are damaged when removing heavy rain. The additive sub-net learns the additive residues which are then added to the heavy rain input to generate a clean image free of rain streaks. The clean estimate $I_1$ of the additive sub-net contains higher frequency information than that of $I_0$. For more details, Fig. \ref{fig:visulization}-(b, c) and Fig. \ref{fig:visulization}-(l, m) show some regions where their structures (vertical lines) are damaged in the outputs $(I_0)$ of the SCA sub-net, and the clean estimate of the additive sub-net $(I_1)$ contains greater details. The multiplicative sub-net learns the multiplicative residues, which selectively extract brighter local features that are not learned effectively in the SCA and additive sub-nets. Such brighter local features are highlighted in the weight map $W_2$ as shown in red boxes in Fig. \ref{fig:visulization}-(j, t). Our attentive blending block adaptively combines the informative features from the three intermediate results to yield a clean and sharp final output image $I$.

As shown in Fig. \ref{fig:visulization}, $W_0$ is active in most image regions, which indicates that the SCA sub-net is the main contributor in generating a clean image. This is reasonable, as the SCA is the unique sub-net with receptive fields large enough to reason about the scene geometries and to remove most rain streaks and haze. On the other hand, it can be observed that $W_1$ is the most active in the darker and detailed regions where rain streaks can be the most prominent in the input images. For this reason, the additive sub-net can effectively help in yielding a final clean image with minimal rain streak artifacts and greater details. We highlight a dark region in $W_1$ where the contributions of $I_1$ can be clearly observed in a yellow box in Fig. \ref{fig:visulization}-(h, r). Finally, $W_2$, as discussed above, shows higher activations in brighter regions (as can be seen in two red boxes in Fig. \ref{fig:visulization}-(j, t)), confirming that the multiplicative residues help modulating the brighter local feature intensities in the final derained output images. Zoom-in Fig. \ref{fig:visulization} for better visualization.

\subsection{Results on the Outdoor-Rain Synthetic Dataset}

\begin{figure*}
\begin{center}
\includegraphics[width=\textwidth,height=\textheight,keepaspectratio]{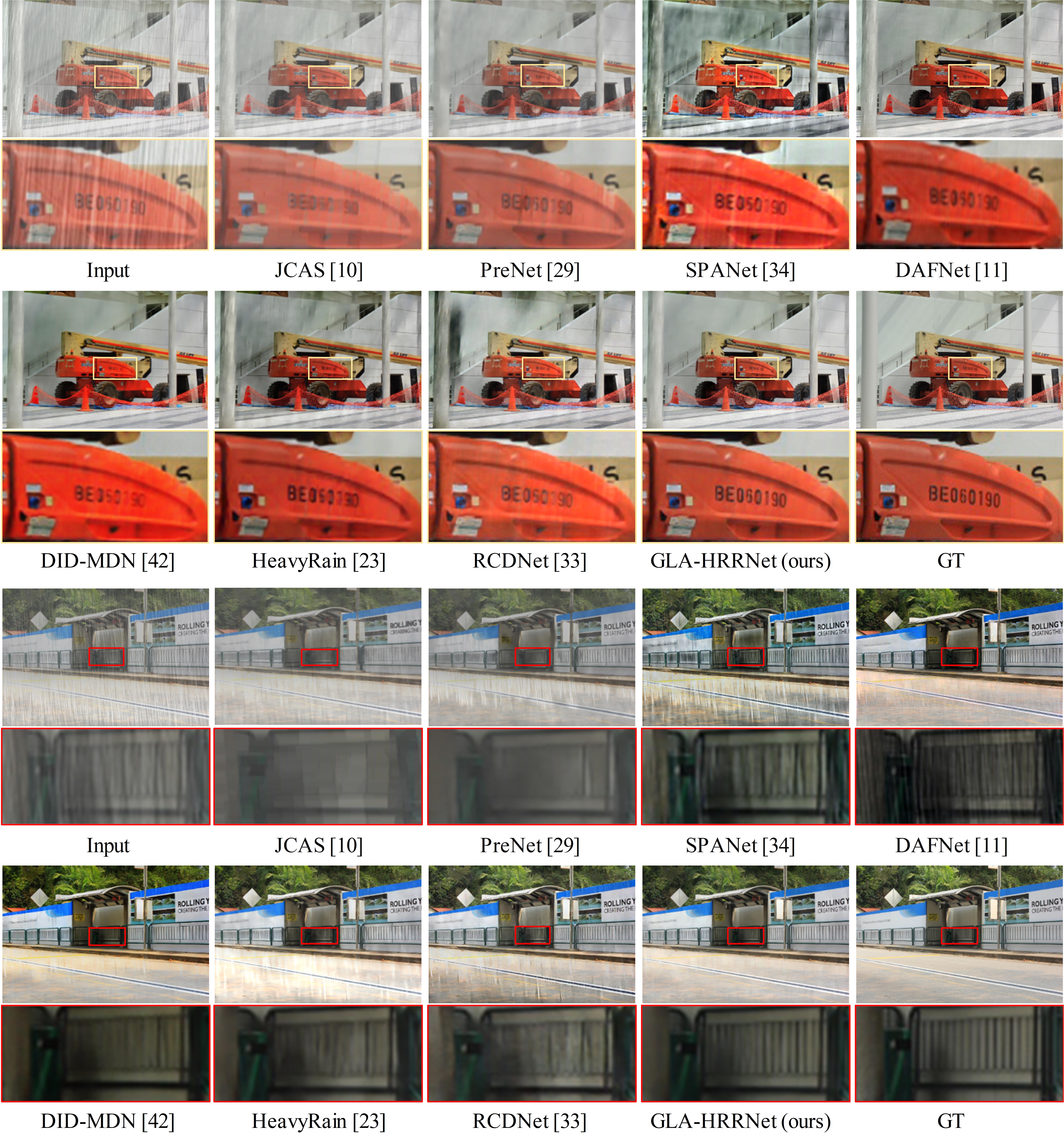}
\end{center}
\caption{Comparison of the proposed model with the SOTA methods on the synthetic Outdoor-Rain dataset \cite{Li_HeavyRain}.}
\label{fig:result-syn}
\end{figure*}

We use the peak signal to noise ratio (PSNR) \cite{PSNR}, and the structural similarity (SSIM) \cite{SSIM} index metrics to compare our GLA-HRRNet with the SOTA methods. Table \ref{tab:result-syn-outdoor} presents a comparison of our model with the following state of the art methods: JCAS \cite{Gu_JCAS_sparse}, RESCAN \cite{Li_RESCAN}, DID-MDN \cite{Zhang_DIDMDN}, SPANet \cite{Wang_SPANet}, PreNet \cite{Ren_PreNet}, UMRL \cite{Yasarla_UMRL}, DAFNet \cite{Hu_DAF}, HeavyRain \cite{Li_HeavyRain}, MSPFN \cite{Jiang_MSPFN} and RCDNet \cite{Wang_RCDNet}. Except the JCAS, all methods in Table \ref{tab:result-syn-outdoor} utilize deep learning-based approaches. It can be observed that the methods trained for plain rain removal \cite{Gu_JCAS_sparse, Ren_PreNet, Wang_SPANet, Li_RESCAN, Jiang_MSPFN} perform very poorly on the heavy rain scenario. Meanwhile, the quantitative performance of our GLA-HRRNet is considerably superior to those of the others, outperforming the SOTA methods by more than \textbf{2dB} in PSNR and \textbf{0.06} in SSIM. Table \ref{tab:result-syn-outdoor} demonstrates the effectiveness of our proposed GLA-HRRNet, which exploits global and local attentions to remove rain and fog entangled in the heavy rains. Moreover, Fig. \ref{fig:result-syn} depicts the qualitative results for our GLA-HRRNet and the SOTA methods. It can be seen that our GLA-HRRNet can effectively remove heavy rains in degraded images and presents very sharp and clean results, while the other methods show unremoved rain streaks, blur, and haze. 

\subsection{Results on the Rain100H Synthetic Dataset}
In addition to the Outdoor-Rain dataset, we compare our GLA-HRRNet with the SOTA method on the Rain100H dataset \cite{Yang_JORDER}. Table \ref{tab:result-syn-100H} presents the performance comparison of our network against the SOTA methods: JCAS \cite{Gu_JCAS_sparse}, RESCAN \cite{Li_RESCAN}, DID-MDN \cite{Zhang_DIDMDN}, UMRL \cite{Yasarla_UMRL}, PreNet \cite{Ren_PreNet}, DAFNet \cite{Hu_DAF}, MSPFN \cite{Jiang_MSPFN}. For the Rain100H dataset, our GLA-HRRNet still outperform all methods under comparison by more than \textbf{1dB} in PSNR and \textbf{0.05} in SSIM. 

\begin{table} 
\setlength{\tabcolsep}{20pt}
\begin{center}
\begin{tabular}{l|l|l}
\hline
\hline
{Method} & PSNR  & SSIM  	\\ \hline
JCAS \cite{Gu_JCAS_sparse}      & 17.69 & 0.683    \\ \hline
RESCAN \cite{Li_RESCAN}         & 26.45 & 0.845    \\ \hline
DID-MDN \cite{Zhang_DIDMDN}     & 25.00 & 0.754    \\ \hline
UMRL \cite{Yasarla_UMRL}  	    & 26.01 & 0.832 	\\ \hline
PreNet \cite{Ren_PreNet}	    & 26.77 & 0.858 	\\ \hline
DAFNet \cite{Hu_DAF}            & 28.44 & 0.874 	\\ \hline
MSPFN \cite{Jiang_MSPFN}	    & 28.66 & 0.860 	\\ \hline
IDAN \cite{Jiang_IADN} 	    & 27.86 & 0.835 	\\ \hline
GLA-HRRNet (ours)    & \textbf{29.88} & \textbf{0.927} \\ \hline
\hline
\end{tabular}
\end{center}
\caption{Quantitative comparison of the proposed model and the SOTA methods on the Rain100H \cite{Yang_JORDER} dataset.}
\label{tab:result-syn-100H}
\end{table}

\subsection{Results on Real Image Datasets}

\begin{figure*}
\begin{center}
\includegraphics[width=\textwidth,height=\textheight,keepaspectratio]{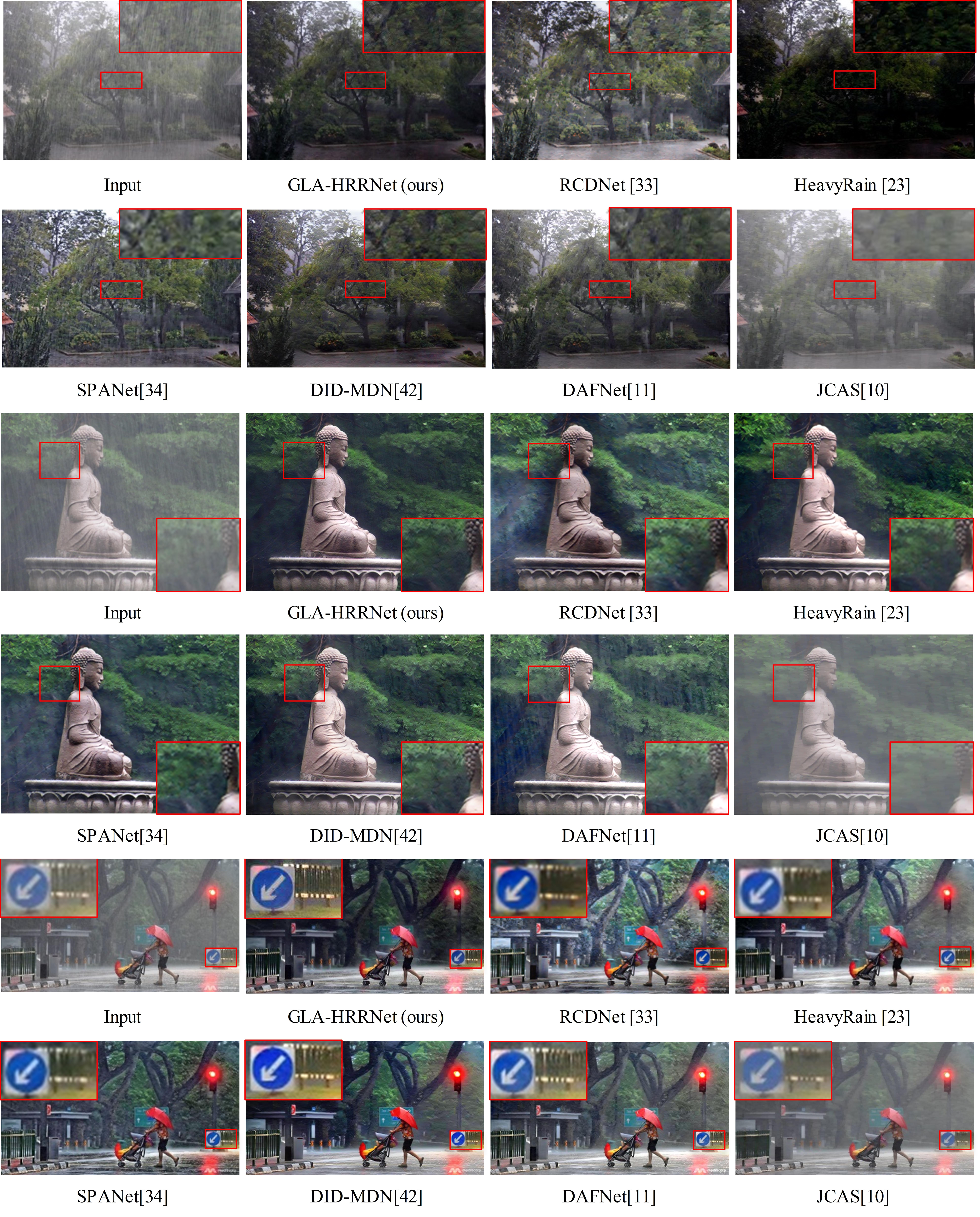}
\end{center}
\caption{The comparison results of proposed methods with SOTAs on real datasets.}
\label{fig:result-real}
\end{figure*}

Fig. \ref{fig:result-real} shows the qualitative comparisons for real heavy rain images among the methods in Table \ref{tab:result-syn-outdoor}. In real scenes, the SOTA methods fail to completely remove heavy rain due to the presence of strong rain accumulation, various shapes, and directions of rain streaks. HeavyRain \cite{Li_HeavyRain} and the DAFNet \cite{Hu_DAF}, which are designed to solve the rain and fog entanglement, do not remove heavy rain efficiently, and show color and contrast that are not realistic. One can easily see the artifacts in the results produced by the SOTA methods. Meanwhile, our method still shows realistic color, cleaner and sharper derained images than the other methods under comparison. Zoom-in Fig. \ref{fig:result-real} for better visualization.

For a more comprehensive comparison, we performed a user study to evaluate the results of SOTA methods on the real image datasets. To do so, we first selected 20 heavy rain images collected from internet with the key word \enquote{heavy rain photo}. For a user study with subjective tests, we invited 20 participants, aged from 20 to 31. Then, we showed the outputs (20 derained images) of the SOTA methods and our GLA-HRRNet to each participant and asked to rate their qualities in perspectives of cleanness and real colorfulness, ranging from 1 to 10. Table \ref{tab:user_study} shows the subjective test results evaluated by the 20 participants for the 20 heavy rain images. As shown in Table \ref{tab:user_study}, our method generates the output images with an average 6.62 score, indicating that our GLA-HRRNet yields better results, which are cleaner and have more realistic colors than other methods.

\begin{table}
\setlength{\tabcolsep}{20pt}
\begin{center}
\begin{tabular}{c|c}
\hline
\hline
Methods & Rating scores (mean \textpm\  std.)   \\ \hline 
DAFNet \cite{Hu_DAF}          & 6.17 \textpm\ 1.53     \\ \hline
DID-MDN \cite{Zhang_DIDMDN}        & 6.10 \textpm\ 1.45    \\ \hline
HeavyRain \cite{Li_HeavyRain}      & 5.89 \textpm\ 1.67    \\ \hline
RCDNet \cite{Wang_RCDNet}         & 5.07 \textpm\ 1.87 \\ \hline
GLA-HRRNet (ours) & \textbf{6.62 \textpm\ 1.64}   \\ \hline
\hline
\end{tabular}
\end{center}
\caption{The user study. Ratings from 1 (bad) to 10 (good) evaluated by participants on the real dataset.}
\label{tab:user_study}
\vspace*{-3mm}
\end{table}

\begin{table}[t]
\setlength{\tabcolsep}{5pt}
\begin{center}
\begin{tabular}{c|c|c}
\hline
\hline
Methods & No. parameters  & Inference times (sec.) \\ \hline
JCAS \cite{Gu_JCAS_sparse}            & x         & 608.9 \\ \hline
DAFNet \cite{Hu_DAF}        & 4.036M    & 0.86 \\ \hline
DID-MDN \cite{Zhang_DIDMDN}        & 66.056M   & 0.07 \\ \hline
HeavyRain \cite{Li_HeavyRain}      & 40.627M   & 0.73 \\ \hline
RCDNet \cite{Wang_RCDNet}        & \textbf{3.166M}    & 1.96 \\ \hline
GLA-HRRNet (ours) & 23.487M   & \textbf{0.02} \\ \hline
\hline
\end{tabular}
\end{center}
\caption{The comparison of the number of parameters and inference time for different methods average on 1500 testing images with size 720x480, in Outdoor-Rain dataset.}
\label{tab:param_infer_time}
\vspace*{-3mm}
\end{table}

\subsection{Complexity comparison with numbers of learnable parameters and inference times}
Table \ref{tab:param_infer_time} compares the computation complexities of rain removal methods in terms of the numbers of parameters and inference times. It can be seen in Table \ref{tab:param_infer_time} that the inference with our proposed GLA-HRRNet is much faster than the other SOTA methods under comparison. For comparisons, the methods are implemented in PyTorch codes (except the JCAS that is implemented in MATLAB) and were tested on a PC platform with two NVIDIA Titan$^{\text{TM}}$ X GPUs and an Intel Core$^{\text{TM}}$ i7-7700K CPU with 64GB RAM.

\subsection{Ablation Study} \label{ablation}
\textbf{Components analysis:} In the ablation study, we evaluated the performance of components in the proposed model. We use the Outdoor-Rain dataset to evaluate six variants of our full model (GLA-HRRNet), listed in Table \ref{tab:ablation-comp}. As shown in Table \ref{tab:ablation-comp}, the SCA sub-net contribution is more significant than the additive and multiplicative sub-nets. Besides, the multiplicative sub-net alone yields the lowest PSNR and SSIM performances. However, the additive and multiplicative sub-nets can bring additional performance improvements in conjunction with the SCA sub-net.

\textbf{Architecture analysis:} The spatial channel attention (SCA) block is the key element of the SCA sub-net. Table \ref{tab:ablation-arch} shows the GLA-HRRNet and its three variants, each of which lacks one of channel attention (denoted as `No CA'), spatial attention (denoted as `No SA') and channel-spatial attention (denoted as `No CA-SA'). For a fair comparison, we add one convolution layer with the kernel size of 3 to the `No CA-SA' variant in order to have similar numbers of parameters.

The PSNR and SSIM results of the three variants and the full model are also shown in Table \ref{tab:ablation-arch}. The `No CA-SA' attains the lowest quantitative result. The `No SA' achieves better performance than the `No CA' and `No CA-SA.' Meanwhile, the full model yields the best result, outperforming the three variants by a large margin.

\begin{table}
\setlength{\tabcolsep}{6pt}
\begin{center}
\begin{tabular}{c|c|c|c|c|c}
\hline
\hline
Network 			& SCA & Add      & Mul            & PSNR & SSIM   \\ \hline
SCA   		& $\surd$&			&				 & 24.46 & 0.848 \\ \hline
Add   			&  		 & $\surd$	&				 & 23.81 & 0.859 \\ \hline
Mul 			&  		 &			&	$\surd$		 & 22.96 & 0.814 \\ \hline
SCA + Add 	& $\surd$& $\surd$	&				 & 24.95 & 0.873 \\ \hline
SCA + Mul 	& $\surd$&			&	$\surd$		 & 24.85 & 0.872 \\ \hline
Add + Mul 	    &  		 & $\surd$	&	$\surd$		 & 24.31 & 0.867 \\ \hline
Full GLA-HRRNet  			& $\surd$& $\surd$ 	&	$\surd$		 & 25.58 & 0.880 \\ \hline
\hline
\end{tabular}
\end{center}
\caption{Ablation studies of the proposed sub-nets.}
\label{tab:ablation-comp}
\end{table}

\begin{table}[t]
\setlength{\tabcolsep}{10pt}
\begin{center}
\begin{tabular}{c|c|c|c|c}
\hline
\hline
Method 	& SA 	 & CA      & PSNR  & SSIM  \\ \hline
No CA   & $\surd$&			& 24.36 & 0.875 \\ \hline
No SA   &  		 & $\surd$	& 25.14 & 0.876 \\ \hline
No CA-SA  &  		 &			& 24.18 & 0.870 \\ \hline
Full GLA-HRRNet 	& $\surd$& $\surd$	& 25.58 & 0.880 \\ \hline
\hline
\end{tabular}
\end{center}
\caption{Ablations studies of our proposed SCA block.}
\label{tab:ablation-arch}
\vspace*{-3mm}
\end{table}

\section{Conclusion}
In this paper, we have proposed an end-to-end network that integrates three sub-nets, called GLA-HRRNet. Our proposed network exploiting global and local attention is capable of effectively removing heavy rain in single images where the strong rain streaks and fog are entangled. The first sub-net, called the \textit{spatial channel attention} (SCA) sub-net, is designed based on a U-net architecture to extract global features that aid in predicting an initial rain-and-fog-free image. The second \textit{additive} sub-net based on our proposed Residual inception modules (RIM) learns to predict additive residues. And finally, the third \textit{multiplicative} sub-net based on our channel-attentive inception modules (CIM) learns informative brighter local features to modulate the local pixel intensities. The final result is produced by adaptively blending our three clean image estimates in our attentive blending block. With the \textit{SCA, additive,} and \textit{multiplicative} sub-nets integrated into our full network, the GLA-HRRNet has shown significantly better performance of heavy rain removal than the recent SOTA methods quantitatively and qualitatively throughout comprehensive experiments for both synthetic and real datasets.

\bibliographystyle{ieee_fullname}
\bibliography{bibliography.bib}

\EOD
\end{document}